\pdfoutput=1

\documentclass[11pt]{article}

\usepackage[]{acl}

\usepackage{times}
\usepackage{latexsym}

\usepackage[T1]{fontenc}

\usepackage[utf8]{inputenc}
\usepackage{microtype}

\usepackage{inconsolata}
\usepackage{booktabs}
\usepackage{graphicx}
\usepackage{multirow}
\usepackage{bbding}

\title{Human-AI Collaborative Essay Scoring: \\A Dual-Process Framework with LLMs}

\author{Changrong Xiao$^1$, \quad Wenxing Ma$^{1}$ \quad Qingping Song$^{2}$ \quad Sean Xin Xu$^1$, \\ \textbf{Kunpeng Zhang}$^{3}$, \quad \textbf{Yufang Wang}$^{4}$, \quad \textbf{Qi Fu}$^{4}$ \\
  $^1$School of Economics and Management, Tsinghua University \\
  $^2$Department of Information Systems, City University of Hong Kong \\
  $^{3}$Department of Decision, Operations \& Information Technologies, University of Maryland \\
  $^{4}$Beijing Xicheng Educational Research Institute\\
  \small{\texttt{xcr21@mails.tsinghua.edu.cn}, \ 
  \texttt{mawx21@mails.tsinghua.edu.cn}, \ 
  \texttt{qisong@cityu.edu.hk}, \ 
  \texttt{xuxin@sem.tsinghua.edu.cn},} \\
  \small{\texttt{kpzhang@umd.edu}, \ 
  \texttt{wangwang7587@163.com}, \ 
  \texttt{lilyhoneypot@163.com}}\\}

\begin{document}
\maketitle
\begin{abstract}

Receiving timely and personalized feedback is essential for second-language learners, especially when human instructors are unavailable. This study explores the effectiveness of Large Language Models (LLMs), including both proprietary and open-source models, for Automated Essay Scoring (AES). Through extensive experiments with public and private datasets, we find that while LLMs do not surpass conventional state-of-the-art (SOTA) grading models in performance, they exhibit notable consistency, generalizability, and explainability. We propose an open-source LLM-based AES system, inspired by the dual-process theory. Our system offers accurate grading and high-quality feedback, at least comparable to that of fine-tuned proprietary LLMs, in addition to its ability to alleviate misgrading. Furthermore, we conduct human-AI co-grading experiments with both novice and expert graders. We find that our system not only automates the grading process but also enhances the performance and efficiency of human graders, particularly for essays where the model has lower confidence. These results highlight the potential of LLMs to facilitate effective human-AI collaboration in the educational context, potentially transforming learning experiences through AI-generated feedback.

\end{abstract}

\section{Introduction}


\begin{figure}[h]
    \centering
    \includegraphics[width=0.5\textwidth]{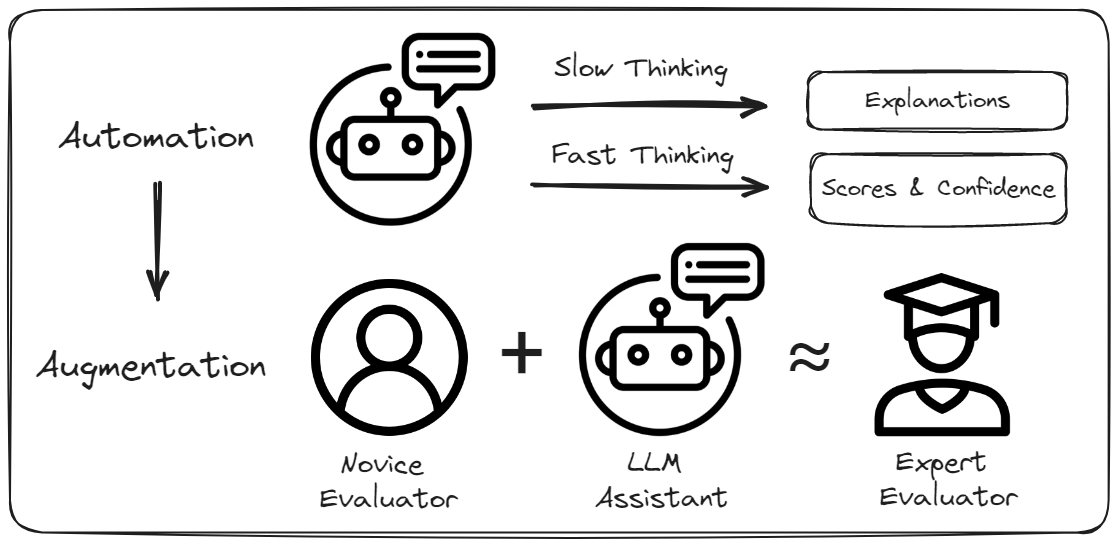}
    \caption{Our study reveals that LLM-based essay scoring systems can not only automate the grading process, but also elevate novice evaluators to the level of experts. }
    \label{page1}
\end{figure}


Writing practice is an essential component of second-language learning. While the provision of timely and reliable feedback poses a considerable challenge for educators in China due to the high student-teacher ratio. This limitation hampers students' academic progress, especially those who are keen on self-directed learning. Automated Essay Scoring (AES) systems provide valuable assistance to students by offering immediate and consistent feedback on their work, and also simplifying the grading process for educators. 


However, implementing AES systems effectively in real-world educational scenarios presents several challenges. First, the diverse range of exercise contexts and the inherent ambiguity in scoring rubrics complicate the ability of traditional models to deliver accurate scores. Second, interviews with high school teachers indicate that despite receiving accurate score predictions, they must still review essays to mitigate potential errors from the models. Consequently, relying exclusively on this system without human supervision is impractical in real-world scenarios. Thus, there is a clear need for AES systems that not only predict scores accurately but also facilitate effective human-AI collaboration. This should be supported by natural language explanations and additional assistive features to enhance usability.


To effectively tackle these challenges, it is crucial to highlight the latest advancements in the field of Natural Language Processing (NLP), particularly focusing on the development of large language models (LLMs). LLMs, such as OpenAI's ChatGPT \footnote{\url{https://chat.openai.com}}, not only showcase impressive capabilities of robust logical reasoning but also exhibit a remarkable ability to comprehend and faithfully follow human instructions \citep{ouyang2022training}. Furthermore, recent studies have highlighted the potential of leveraging LLMs in AES tasks \citep{mizumoto2023exploring, yancey-etal-2023-rating, naismith-etal-2023-automated}.

In this study, we explore the potential of proprietary and open-source LLMs such as GPT-3.5, GPT-4, and LLaMA3 for AES tasks. We conducted extensive experiments with public essay-scoring datasets as well as a private collection of student essays to assess the zero-shot and few-shot performance of these models. Additionally, we enhanced their effectiveness through supervised fine-tuning (SFT). Drawing inspiration from the dual-process Theory, we developed an AES system based on LLaMA3 that matches the grading accuracy and feedback quality of fine-tuned LLaMA3. Our human-LLM co-grading experiment further revealed that this system significantly improves the performance and efficiency of both novice and expert graders, offering valuable insights into the educational impacts and potential for effective human-AI collaboration. Overall, our study contributes three major advancements to the field:


\begin{itemize}
    \item We pioneer the exploration of LLMs' capabilities as AES systems, especially in complex scenarios featuring tailored grading criteria. Leveraging dual-process theory, our novel AES framework demonstrates remarkable accuracy, efficiency, and explainability.
    \item We introduce an extensive essay-scoring dataset, which includes 13,372 essays written by Chinese high school students. These essays are evaluated with multi-dimensional scores by expert educators. This dataset significantly enhances the resources available for AI in Education (AIEd)\footnote{Codes and resources can be found in \url{https://github.com/Xiaochr/LLM-AES}}. 
    \item Our findings from the human-LLM co-grading task highlight the potential of LLM-generated feedback to elevate the proficiency of individuals with limited domain expertise to a level akin to that of experts. Additionally, it enhances the efficiency and robustness of human graders by integrating model confidence scores and explanations. These insights set the stage for future investigation into human-AI collaboration and AI-assisted learning within educational contexts. 
\end{itemize}


\section{Related Work}

\subsection{Automated Essay Scoring (AES)}



\paragraph{Traditional Methods} Automated Essay Scoring (AES) stands as a pivotal research area at the intersection of NLP and education. Traditional AES methods are usually regression-based or classification-based machine learning models \citep{sultan2016fast, mathias2018thank, mathias2018asap++, salim2019automated} trained with textual features extracted from the target essays. With the advancement of deep learning, AES has witnessed the integration of advanced techniques such as convolutional neural networks (CNNs) \citep{dong2016automatic}, long short-term memory networks (LSTMs) \citep{taghipour2016neural}, and also pre-trained language models \citep{rodriguez2019language, lun2020multiple}. These innovations have led to more precise score predictions, and state-of-the-art methods are primarily based on Bidirectional Encoder Representations from Transformers (BERT) \citep{yang2020enhancing, wang2022use, boquio-naval-jr-2024-beyond-canonical}.


\paragraph{LLM Applications in AES} Recent studies have explored The potential of leveraging the capabilities of modern LLMs in AES tasks. \citet{mizumoto2023exploring} provided ChatGPT with specific IELTS scoring rubrics for essay evaluation but found limited improvements when incorporating GPT scores into the regression model. In a different approach, \citet{yancey-etal-2023-rating} used GPT-4's few-shot capabilities to predict Common European Framework of Reference for Languages (CEFR) levels for short essays written by second-language learners. However, the Quadratic Weighted Kappa (QWK) scores still did not surpass those achieved by the XGBoost baseline model or human annotators. Similarly, \citet{han2023fabric, stahl2024exploring} introduced prompting frameworks that did not outperform the conventional baselines. 


\subsection{AI-Assisted Decision Making}


Researchers have extensively investigated human-AI teams, in which AI supports the decision-making process by providing recommendations or suggestions, while the human remains responsible for the final decision \citep{van2019six}. The objective of such human-AI collaboration is to achieve complementary performance, where the combined team performance exceeds that of either party operating independently \citep{bansal2021does}. To realize this, it is crucial to design an AI-assisted decision-making process that allows humans to effectively monitor and counteract any unpredictable or undesirable behavior exhibited by AI models \citep{eigner2024determinants}. This design aims to leverage the strengths of both humans and AI to enhance overall performance \citep{holstein2022designing}. To our knowledge, no studies have yet investigated AES systems from this angle of collaborative co-grading.

\subsection{Dual-Process Theory}

Recent studies have developed architectures that imitate human cognitive processes to enhance the capabilities of LLMs, particularly in reasoning and planning \citep{benfeghoul2024doubt}. According to dual-process theory in psychology \citep{wason1974dual, kahneman2011thinking}, human cognition operates via two distinct systems: System 1 involves rapid, intuitive "Fast Thinking", while System 2 entails conscious and deliberate "Slow Thinking" processes. LLM architectures inspired by this theory have been implemented in complex interactive tasks \citep{lin2024swiftsage, tian2023duma}, aiming to mitigate issues like social biases \citep{kamruzzaman2024prompting} and hallucination \citep{bellini2023dual}. These adaptations have demonstrated improved performances in various areas.

\section{Data}

\paragraph{ASAP dataset} The Automated Student Assessment Prize (ASAP\footnote{\url{https://www.kaggle.com/c/asap-aes.}}) dataset, stands as one of the most commonly used publicly accessible resources Automated Essay Scoring (AES) tasks. This comprehensive dataset comprises a total of $12,978$ essays, encompassing responses to $8$ distinct prompts. Each essay has been evaluated and scored by human annotators. Essay sets are also accompanied by detailed scoring rubrics, each tailored with unique scoring guidelines and score ranges. These intricacies are essential as they cater to the multifaceted requirements and diverse scenarios of AES.

\paragraph{Our Chinese Student English Essay (CSEE) dataset} We have developed a novel English essay scoring dataset specifically designed for AES tasks. The dataset was carefully curated in collaboration with $29$ high schools in China, encompassing a total of $13,372$ student essays responding to two distinct prompts used in final exams. The evaluation of these essays was carried out by highly experienced English teachers following the scoring guidelines of the Chinese National College Entrance Examination (Table \ref{rubrics}). Scoring was comprehensively assessed across three critical dimensions: Content, Language, and Structure, with an Overall Score ranging from 0 to 20. More descriptions of the two datasets are presented in Appendix \ref{datasets}.


\section{Methods}

\begin{figure*}[h]
    \centering
    \includegraphics[width=0.95\textwidth]{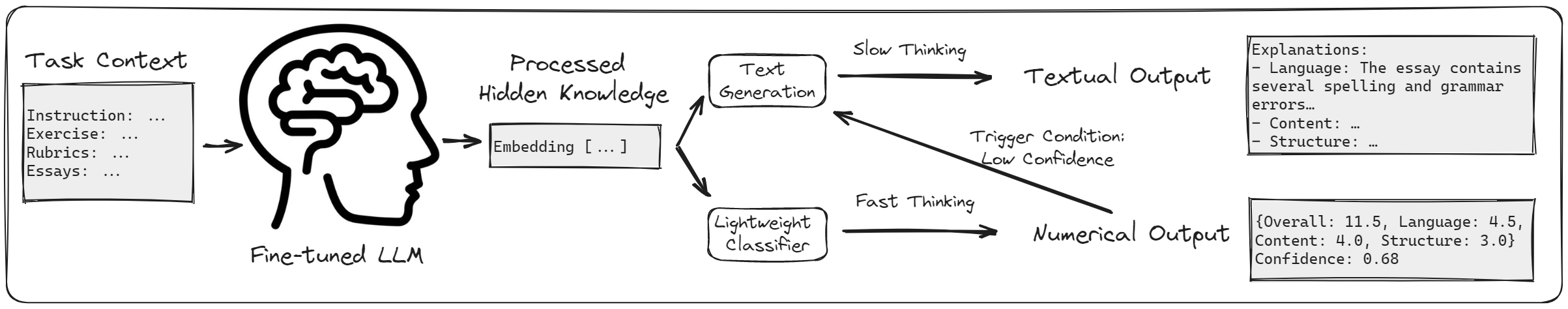}
    \caption{Our proposed Fast and Slow Thinking AES framework. }
    \label{fast_slow}
\end{figure*}


In this section, we present the details of the models used in this study, including traditional AES baselines, LLM-based approaches, and our proposed Fast and Slow Thinking AES framework. 

\subsection{Traditional Essay Scoring Baselines}

\paragraph{BERT Classifier} Similar to the model used in \citet{yang2020enhancing, han2023fabric}'s work, we implemented a simple yet effective baseline model for score prediction based on BERT. This model integrated a fully connected prediction layer following the BERT output, and the BERT parameters remained unfrozen during training. Both the BERT model and the prediction layer were jointly trained on the training essay set (details in Appendix \ref{bert_appendix}). 

\paragraph{SOTA baselines} We also incorporate models such as \textit{R$^2$BERT} \citep{yang2020enhancing} and \textit{Tran-BERT-MS-ML-R} \citep{wang2022use}, which represent the highest levels of performance in the ASAP AES task. These models serve as the high-level benchmarks against which we evaluate the performance of our LLM-based models.

\subsection{Prompting LLMs}


We considered various prompting strategies including with or without detailed rubrics context, zero-shot or few-shot settings. An illustrative example of a prompt and its corresponding model-generated output can be found in Table \ref{example} in the Appendices. 

\paragraph{GPT-4, zero-shot, without rubrics} In this setting, we simply provide the prompt and the target essay to GPT-4. The model then evaluates the essay and assigns a score based on its comprehension within the specified score range.

\paragraph{GPT-4, zero-shot, with rubrics} Alongside the prompt and the target essay, we also provide GPT-4 with explicit scoring rubrics, guiding its evaluation.

\paragraph{GPT-4, few-shot, with rubrics} In addition to the zero-shot settings, the few-shot prompts include sample essays and their corresponding scores. This assists GPT-4 in understanding the latent scoring patterns. With the given prompt, target essay, scoring rubrics, and a set of $k$ essay examples, GPT-4 provides an appropriate score reflecting this enriched context. See Appendix \ref{append_llm_methods} for details. 

In all these configurations, we adopted the Chain-of-Thought (CoT) \citep{wei2022chain} strategy. This approach instructed the LLM to analyze and explain the provided materials before making final score determinations. Studies \citep{lampinen-etal-2022-language, zhou2023leasttomost, li-etal-2023-making} have shown that this structured approach significantly enhances the capabilities of the LLM, optimizing performance in tasks that require inference and reasoning. 

\subsection{Fine-tuning LLMs}

We conducted additional investigations into the effectiveness of supervised fine-tuning methods. Given that the ASAP and our CSEE dataset only include scores without expert explanations, we augmented these original datasets with explanations generated by GPT-4. To guide the explanation generation process, we provided GPT-4 with a few expert-curated explanations and a structured template. By organizing the data into an instructional format, we created fine-tuning inputs that enable the LLMs to not only generate accurate scores but also provide high-quality feedback.

We first fine-tuned OpenAI's \textit{GPT-3.5-turbo}, one of the best-performing LLMs. However, due to the proprietary nature of GPT-3.5 and considerations such as data privacy, training and inference costs, and flexibility in fine-tuning, we also fine-tune an LLaMA3-8B \footnote{\url{https://llama.meta.com/llama3/}} model with both original and augmented datasets. This recent open-source model mitigates these concerns and has remarkable capabilities, making it a more practical choice for use in educational scenarios.


\subsection{Our Proposed Method}

As previously mentioned, score prediction and explanation generation are distinct but interrelated tasks within the context of AES. Explanation generation, which covers the evaluation of content, language, and structure, necessitates deliberate and meticulous reasoning. On the other hand, score prediction can either be a swift process based on intuition and experiences or concluded after step-by-step inference. These features align with the idea of dual-process theory. Consequently, we have designed an integrated system that includes separate modules for each task: the Fast Module for quick score prediction and the Slow Module for detailed explanation generation. The framework of our proposed AES system is shown in Figure \ref{fast_slow}.

\paragraph{Slow Module: Fine-tuned LLM} The Slow Module forms the core of our AES system, capable of analyzing essays in depth, providing evidence based on specific rubrics, and deriving appropriate scores. This detailed process is time-intensive but yields valuable natural language reasoning that informs the final grading decision. In this study, we implemented the fine-tuned LLaMA3-8B as the Slow Module. It is worth noting that this module is interchangeable with any other qualified LLM, demonstrating the flexibility of our framework.

\paragraph{Fast Module: Lightweight Classifier} In many cases, swift score prediction is preferable to detailed reasoning. To reduce the time and computational cost associated with generating detailed explanations, we introduced a simple fully connected layer as a bypass before the initiation of text generation by the Slow Module. By using only the embeddings of the input text, the Fast Module not only conserves resources but also leverages the latent knowledge acquired during the fine-tuning of the Slow Module, aligning with the 'intuitive' facet of Fast Thinking. 

When to switch from the Fast to Slow Thinking module is one of the challenges in the design of dual-process LLM. Previous frameworks employed heuristic rules or error feedback as the triggering criteria \citep{lin2024swiftsage, tian2023duma}, which might be impractical in real-world cases. Our Fast module also calculates the probabilities of each possible output score, which we standardize and treat as confidence scores. Predictions with low confidence are considered unreliable, triggering the Slow Module for self-reflection, or passing to external judges (either human or AI). This design aims to enhance essay scoring accuracy and foster effective human-AI collaboration, potentially elevating the complementary team performance.

For training, we first fine-tune the Slow Module using our explanation-augmented dataset. Subsequently, we employ the Slow Module to derive input embeddings, which, paired with the rated scores, are used to train the Fast Classifier from scratch. During inference, essay inputs initially pass through the fine-tuned LLM and are transformed into the embedding format. They are then processed by the Fast Module to quickly derive scores. The Slow Module is activated only when prediction confidence is low or based on specific additional requirements.




\section{Experimental Results}

\begin{table*}[h]
\centering
\small
\caption{Comparison of QWK scores for LLM-based methods and the baselines under the ASAP dataset. The "E." column indicates whether the model output includes natural language explanations alongside the predicted scores. }
\label{asap_qwk}
\begin{tabular}{@{}l|l|ccccccccc@{}}
\toprule
                                & E. &  Set 1  & Set 2  & Set 3  & Set 4  & Set 5  & Set 6  & Set 7  & Set 8 & Avg. \\ \midrule
BERT Classifier               & \XSolidBrush  & 0.6486 & 0.6284 & 0.7327 & 0.7669 & 0.7432 & 0.6810 & 0.7165 & 0.4624 & 0.6725 \\ 
Tran-BERT-MS-ML-R              & \XSolidBrush   & 0.8340 & 0.7160 & 0.7140 & 0.8120 & 0.8130 & 0.8360 & 0.8390 & 0.7660 & 0.7910 \\ 
R$^2$BERT                & \XSolidBrush   & 0.8170 & 0.7190 & 0.6980 & 0.8450 & 0.8410 & 0.8470 & 0.8390 & 0.7440 & 0.7940 \\ 
\midrule
GPT-4, zero-shot, w/o rubrics  & \Checkmark  & 0.0423 & 0.4017 & 0.2805 & 0.5571 & 0.3659 & 0.5021 & 0.0809 &  0.4188 & 0.3312 \\
GPT-4, zero-shot, with rubrics & \Checkmark & 0.0715 & 0.3003 & 0.3661 & 0.6266 & 0.5227 & 0.3448 & 0.1101 & 0.4072 & 0.3437  \\
GPT-4, few-shot, with rubrics  & \Checkmark  & 0.2801 & 0.3376 & 0.3308 & 0.7839 & 0.6226 & 0.7284 & 0.2570 & 0.4541 & 0.4743 \\ 
\midrule
Fine-tuned GPT-3.5  & \XSolidBrush  & 0.7406 & 0.6183 & 0.7041 & \textbf{0.8593} & 0.7959 & \textbf{0.8480} & \textbf{0.7271} & \textbf{0.6135} & \textbf{0.7384} \\
Fine-tuned LLaMA3           & \XSolidBrush   & 0.7137 & \textbf{0.6696} & 0.6558 & 0.7712 & 0.7452 & 0.7489 & 0.6938 & 0.2952 & 0.6617 \\ 
\midrule
Ours                & \Checkmark   & \textbf{0.7612} & 0.6517 & \textbf{0.7238} & 0.8093 & \textbf{0.8118} & 0.7764 & 0.7071 & 0.4885 & 0.7162 \\ 
\quad Fast module                &  \XSolidBrush  & 0.7580 & 0.6395 & 0.7228 & 0.7995 & 0.8023 & 0.7753 & 0.7157 & 0.5075 & 0.7151 \\ 
\quad Slow module           & \Checkmark   & 0.6048 & 0.5621 & 0.5700 & 0.6992 & 0.6774 & 0.5943 & 0.5772 & 0.2677 & 0.5691 \\ 
\bottomrule
\end{tabular}
\end{table*}

\begin{table*}[h]
\centering
\small
\caption{Comparison of QWK scores for LLM-based methods and the baselines under our CSEE dataset. The "E." column indicates whether the model output includes natural language explanations alongside the predicted scores. }
\label{stu_qwk}
\begin{tabular}{@{}l|l|cccc@{}}
\toprule
                                & E. & Overall & Content & Language & Structure \\ \midrule
BERT Classifier                   &\XSolidBrush & \textbf{0.7674}  & 0.7312 & 0.7203 & 0.6650    \\ \midrule
GPT-4, zero-shot, w/o rubrics  & \Checkmark & 0.4688  & 0.4412 & 0.3081 &  0.5757  \\
GPT-4, zero-shot, with rubrics   & \Checkmark & 0.5344  & 0.5391 & 0.4660 & 0.4256    \\
GPT-4, few-shot, with rubrics   & \Checkmark &  0.6729 & 0.6484 & 0.6278 & 0.4661    \\ \midrule
Fine-tuned GPT-3.5   &\XSolidBrush & 0.7532  & 0.7241 & \textbf{0.7513} & 0.6576    \\
Fine-tuned LLaMA3              &\XSolidBrush & 0.7544 & 0.7321 & 0.7084 & 0.6461  \\ 
\midrule
Ours        & \Checkmark   & 0.7634 & \textbf{0.7347} & 0.7192 & \textbf{0.6656}  \\ 
\quad Fast module        &  \XSolidBrush  & 0.7364 & 0.7272 & 0.7072 & 0.6627  \\ 
\quad Slow module      & \Checkmark  & 0.7310 & 0.6810 & 0.6990 & 0.6412  \\ 
\bottomrule
\end{tabular}
\end{table*}

\subsection{Performance of LLM-based Methods}


We conducted experiments across all eight subsets of the ASAP dataset using both the LLM-based methods and baseline approaches. We adopted Cohen's Quadratic Weighted Kappa (QWK) as our primary evaluation metric, which is the most widely recognized automatic metric in AES tasks \citep{ramesh2022automated}. A higher QWK value indicates a greater degree of agreement between the predicted score and the ground truth. For methods requiring a training dataset, we divided the data for each subset using an 80:20 split ratio between training and testing.

Our extensive experiments, as detailed in Table \ref{asap_qwk}, revealed that despite using carefully curated prompts and providing detailed context, the zero-shot and few-shot capabilities of GPT-4 did not yield high QWK scores on the ASAP dataset. In zero-shot scenarios, GPT-4's performance was notably low, with some subsets scoring nearly as poorly as random guessing. For instance, Set 1 recorded a QWK of 0.0423 and Set 7 a QWK of 0.0809. This underperformance may be due to the broad scoring ranges and complex rubrics in ASAP, suggesting that even advanced LLMs like GPT-4 may struggle to fully comprehend and adhere to complicated human instructions. In few-shot settings, although there was an improvement in scoring performance, particularly for Sets 4-6, GPT-4 still significantly lagged behind SOTA grading methods. This is consistent with findings from recent studies that utilize LLMs for essay scoring.

When fine-tuned with the training dataset, the LLMs demonstrated significantly improved performance compared to the zero-shot and few-shot results, with QWK scores generally exceeding 0.7. However, these fine-tuned LLMs still did not surpass traditional SOTA methods. Within our framework, the performance of the fine-tuned open-source LLaMA3-8B was comparable to that of fine-tuned proprietary models. Even simple supervised fine-tuning (SFT) of LLaMA3 achieved notable results, suggesting that open-source LLMs might be a cost-effective choice for AES tasks. The findings from our CSEE dataset (see Table \ref{stu_qwk}) align with those on the ASAP dataset, indicating that our framework predicts reliable scores across content, language, and structure dimensions.


Although LLMs do not match traditional methods in terms of scoring accuracy, they excel at generating detailed explanations, benefiting both educators and students. Notably, when trained to produce both scores and explanations in a single output (our proposed Slow Module), LLaMA3-8B experienced a performance drop in grading accuracy. This decrease may be attributed to the model's optimization process, where numerical score values are treated similarly to textual data in the output, leading to suboptimal accuracy. In our Fast and Slow Thinking framework, however, separating numerical from textual outputs and integrating them based on a trigger condition improved the QWK scores, enhancing collaborative performance. Additionally, we evaluated the quality of explanations generated by our AES system against those produced by GPT-4. Through a comparison competition among crowdsourced workers, analyzing 20 sets of paired essay grading explanations, our system achieved a win rate of 35\%, a tie rate of 40\%, and a loss rate of 25\%. These results demonstrate that our explanations are of high quality and comparable to those generated by GPT-4.

\subsection{Further Analyses}

\paragraph{Consistency} To assess the consistency of scores predicted by LLM-based methods, we conducted the same experiment three times, each with the \textit{temperature} parameter of the LLMs set to $0$. We observed that over $80\%$ of the ratings remained unchanged across these trials, indicating a high level of consistency. We then computed the average of these three values to determine the final results.


\paragraph{Generalizability} The eight subsets of the ASAP dataset, featuring diverse scoring criteria and ranges, serve as an excellent framework for evaluating the generalization capabilities of models. For methods such as fine-tuning and traditional baselines that require training data, we first trained the models on one subset and then assessed their performance across the remaining datasets. For example, we trained on Set 1 and tested on Sets 2-8, keeping the model weights fixed. We selected fine-tuned GPT-3.5 and the BERT Classifier to represent LLM-based and traditional methods, respectively. As detailed in Table \ref{gen}, our fine-tuned GPT-3.5 generally outperformed the BERT classifier, although there were instances of underperformance, notably when trained on Set 4 and tested on Sets 1 and 7. The BERT classifier showed particularly weak generalization when trained on Sets 7 and 8, performing close to random guessing.

\begin{figure}[h]
    \centering
    \includegraphics[width=0.48\textwidth]{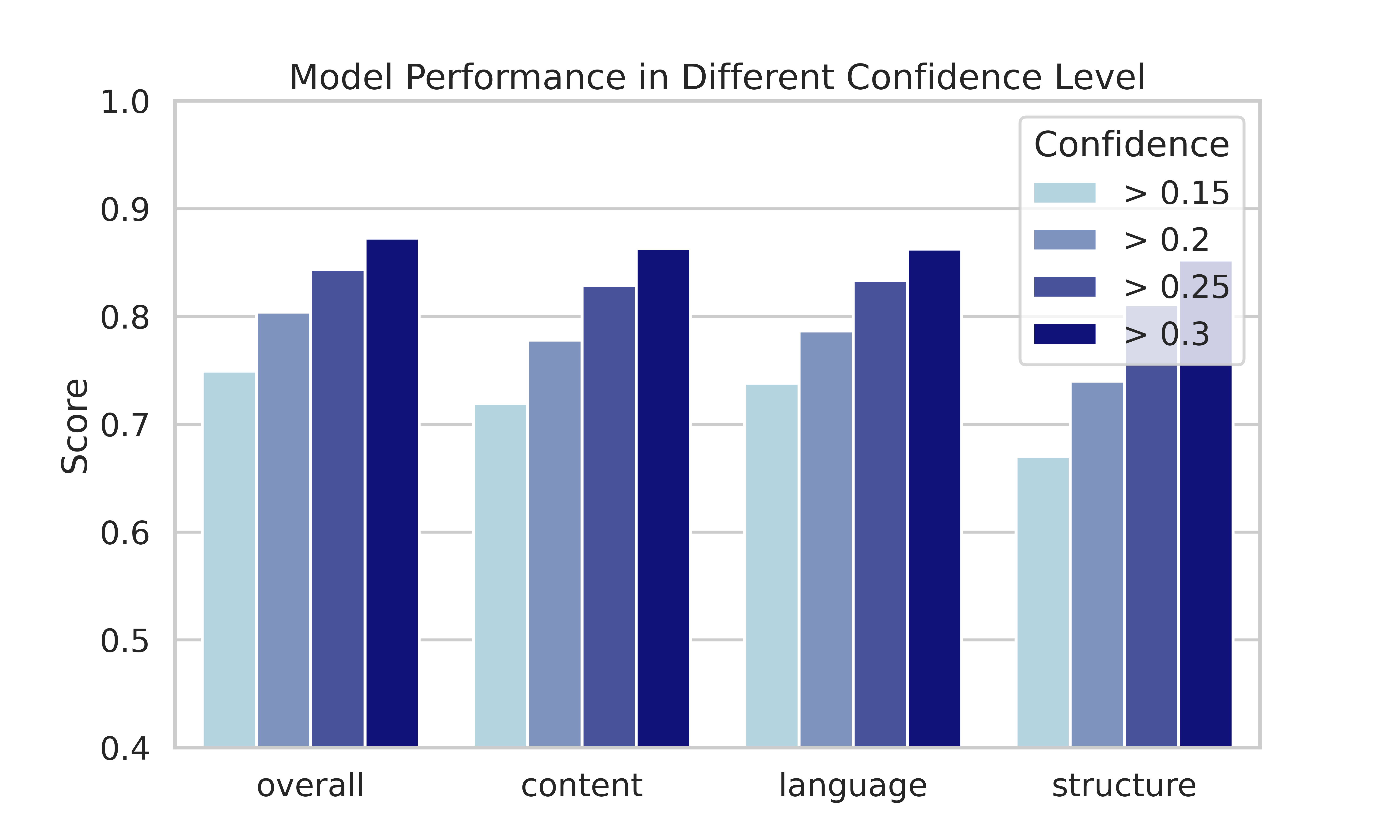}
    \caption{QWK scores of our Fast module in different confidence levels. }
    \label{conf_img}
\end{figure}

\paragraph{Prediction Confidence and Self-Reflection} To assess the reliability of confidence scores, we segmented the test samples based on the output confidence and observed a strong correlation between these scores and model performance in Figure \ref{conf_img}. The trigger condition for switching from the Fast to the Slow Module is set when the confidence score falls below 0.2. Although the Slow Module generally exhibits lower performance compared to the Fast Module, the overall performance of the integrated system improved. This enhancement suggests that employing detailed reasoning for cases with low confidence is an effective grading strategy.

\paragraph{Time Efficiency} Training the Slow Module for each epoch with our explanation-augmented dataset requires around 2 hours using an RTX 4090 24GB GPU, and the inference process consumes about 10 GPU hours. Meanwhile, training the Fast Module takes less than 0.5 hours, and scoring predictions are completed in just 0.2 hours. Our proposed framework, which incorporates a confidence trigger condition, offers an effective trade-off by enhancing both accuracy and efficiency.

\section{Human-AI Co-Grading Experiment}



Given that the AES system not only provides score predictions but also functions as a teammate to educators, we further explore the effectiveness of our proposed system in assisting human grading.

\subsection{Experiment Design}

To investigate the performance of human-only, AI-only, and human-AI team collaboration, we conducted a two-stage within-group experiment. We randomly selected 50 essays from the test set of our CSEE dataset, all on the same topic. We recruited 10 college students from a Normal University in Beijing, who are prospective high school teachers with no current grading experience, to serve as novice evaluators. Additionally, 5 experienced high school English teachers participated as expert evaluators. Initially, all evaluators graded the essays independently using standard rubrics. Subsequently, they were provided with the scores, prediction confidence levels, and explanations generated by our AES system and had the option to revise their initial scores based on this augmented information. To gather feedback on the process, we distributed questionnaires where evaluators rated their experience on a 5-point Likert scale, with higher scores indicating better-perceived performance.



In short, we mainly focus on the following research questions:
\begin{itemize}
    \item Can novice and expert human evaluators achieve complementary performance in terms of accuracy and efficiency using the proposed AES system and collaborative workflow?
    \item Does the design of prediction confidence and explanation generation contribute to performance improvements?
\end{itemize}

\subsection{Results}

\begin{figure}[h]
    \centering
    \includegraphics[width=0.4\textwidth]{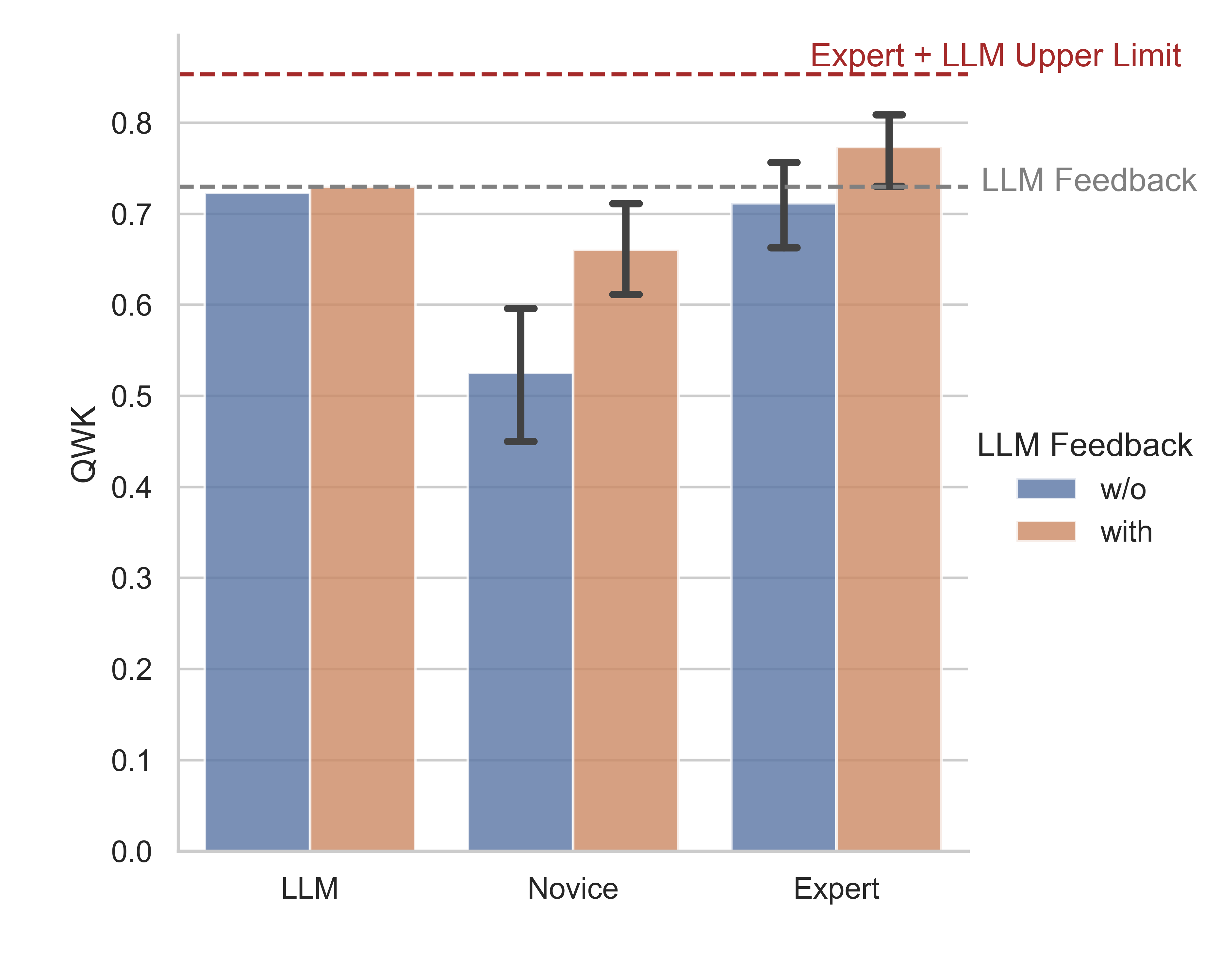}
    \caption{QWK of the overall score in LLM-assisted co-grading experiment for the novice and expert evaluators. The blue bar and orange bar of the LLM column indicate the performance of our Fast module and the integrated system respectively. }
    \label{human_eval_res}
\end{figure}

\paragraph{Feedback generated by LLM elevates novice evaluators to expert level.} As depicted in Figure \ref{human_eval_res} and Table \ref{ttest}, our findings reveal that novice graders, with the assistance of LLM-generated feedback (including both scores and explanations), achieved a significant improvement in performance. Their average QWK improved from 0.5256 to 0.6609, with a p-value of less than 0.01. Furthermore, when comparing the performance of LLM-assisted novice evaluators (mean QWK of 0.6609) to that of expert graders (mean QWK of 0.7117), no statistical difference was found between the two groups (p-value = 0.27). This indicates that with LLM support, novice evaluators achieved a level of grading proficiency comparable to that of experienced experts. Similar trends were observed in the scores for content, language, and structure, with detailed results presented in Table \ref{detail_human_eval_res}.

\begin{table}[h]
\centering
\small
\caption{$t$-test of QWK scores for different experimental groups. \textit{Diff.} means the difference of means between the two groups of QWK scores. }
\label{ttest}
\begin{tabular}{@{}lccc@{}}
\toprule
                     & Diff.   & $t$ statistic & $p$-value \\ \midrule
Expert vs. Novice     & \textbf{0.1860***}  & 3.2152       & \textbf{0.0068}  \\
Novice+LLM vs. Novice & \textbf{0.1353***}  & 2.8882      & \textbf{0.0098}  \\
Expert+LLM vs. Expert & 0.0617  & 1.7128      & 0.1251  \\
Novice+LLM vs. Expert & -0.0508 & -1.1566     & 0.2682  \\ \bottomrule
\end{tabular}
\end{table}

\paragraph{Feedback generated by LLM boosts expert efficiency and consistency.} The integration of LLM-generated feedback into the expert grading process led to an increase in the average QWK from 0.7117 to 0.7734, which also surpassed the performance of AES systems (a QWK of 0.7302) for these essay samples, thereby achieving superior complementary performance (where the Human-AI team outperforms both individual human and AI). Although this improvement is not statistically significant ($p$-value = 0.13), the benefits of LLM augmentation for experts were evident in several other aspects. According to self-report questionnaires (refer to Table \ref{expert_q}), experts required less time to complete grading tasks when assisted by the LLM. Furthermore, a reduction in the standard deviation of expert ratings was observed, indicating a higher level of consensus among experts. This suggests that LLM-generated feedback leads to more consistent evaluations of student essays. Experienced domain experts also commended the accuracy and practicality of the LLM-generated feedback, particularly praising the prediction confidence mechanism which alerted them to scrutinize more challenging cases. These findings highlight the potential to augment the human grading process with our AES system in real-world educational environments.

\begin{table}[h]
\centering
\small
\caption{Experts' feedback after grading student essays with the support of the LLM-based system. }
\label{expert_q}
\begin{tabular}{@{}p{0.35\textwidth}c@{}}
\toprule
                                                                                & Score \\ \midrule
Perceived accuracy of LLM overall score                                                 & \textbf{4.3}/5          \\
Perceived accuracy of LLM content score                                               & \textbf{4.0}/5          \\
Perceived accuracy of LLM language score                                              & 3.9/5          \\
Perceived accuracy of LLM structure score                                             & 3.8/5          \\ \midrule
Helpfulness of the predicted scores                                                    & \textbf{4.6}/5          \\
Helpfulness of the confidence scores                                                    & \textbf{4.8}/5          \\
Helpfulness of LLM explanations                                             & \textbf{4.7}/5          \\
Efficiency of LLM assistance                                             & \textbf{4.4}/5          \\
Willingness to use our AES system & \textbf{4.3}/5          \\ \bottomrule
\end{tabular}
\end{table}


\paragraph{The Importance of Prediction Confidence and Explanations} We previously assessed the reliability of prediction confidence from our Fast Module and noted a modest improvement in model performance after self-reflection by the Slow Module (as shown in the LLM column of Figure \ref{human_eval_res}). In the context of human-AI collaboration, we focused on cases where the predicted scores presented to human evaluators were of low confidence (below 0.2). We observed that the overall QWK scores for expert and novice evaluators were 0.6809 and 0.5680. These QWK values, lower than the average human performances, suggest that these essays are inherently challenging to grade, even for humans. However, human performances exceeded that of the LLM Slow Module’s 0.5478 QWK, achieving complementary team performance. These findings support a practical, intuitive LLM-assisted decision-making workflow: the model manages routine cases with high confidence and minimal human intervention, while low-confidence cases are presented to human collaborators for in-depth analysis and final decision-making.

\section{Conclusion}


In this study, we explored the capabilities of LLMs within AES systems. With detailed contexts, clear rubrics, and high-quality examples, GPT-4 demonstrated satisfactory performance, consistency, and generalizability. Further accuracy enhancements were achieved through supervised fine-tuning using task-specific instruction datasets, bringing LLM performance close to conventional SOTA methods. To leverage the LLMs' ability to generate natural language explanations along with predicted scores, we introduced an open-source Fast and Slow Thinking AES framework. This framework not only matches the quality of proprietary models but also offers greater efficiency.


Our research extended into human-AI co-grading experiments within this new framework. A notable finding was that LLMs not only automated the grading process but also augmented the grading skills of human evaluators. Novice graders, with support from our AES framework, reached accuracy levels comparable to those of experienced graders, while expert graders showed gains in efficiency and consistency. The collaboration between humans and AI particularly enhanced performance in handling low-confidence cases, demonstrating a significant synergy that approached the upper limits of team performance. These results highlight the transformative potential of AI-assisted and human-centered decision-making workflows, especially in elevating those with limited domain knowledge to expert-level proficiency. This study illuminates promising future directions for human-AI collaboration and underscores the evolving role of AI in educational contexts.

\section*{Limitations}

This study has certain limitations. Firstly, although our CSEE dataset includes a substantial number of student essays, these essays originate from only two final exams designed for high school English learners in China. This raises concerns about the robustness of our proposed AES system when applied to a broader range of topics and diverse student demographics. Secondly, our human-AI collaboration experiment, while indicative of promising directions for future human-AI co-grading tasks, is a pilot study that yields general results. Further experiments are necessary to thoroughly explore the mechanisms of complementary team performance, such as identifying circumstances under which humans are likely to recognize and correct their errors following AI feedback, or instances where unreliable AI feedback could potentially mislead them. A deeper understanding of these collaboration mechanisms will enable researchers to develop AES systems that offer more effective support to educators.

\section*{Ethical Considerations}

We secured Institutional Review Board (IRB) approval for both the data collection and the human-AI co-grading experiment (details provided in the online materials). In our CSEE dataset, all personal information concerning the students has been anonymized to safeguard their privacy. The dataset comprises solely of essays and the corresponding scores, omitting any additional information that might raise ethical concerns. However, details of the data annotation process remain undisclosed to us, including the number of teachers involved in the scoring and the level of inter-annotator agreement among them. We have also obtained explicit consent to use the data exclusively for research purposes from both teachers and students.

\bibliography{anthology,custom}

\appendix

\newpage

\section{Datasets}
\label{datasets}

The details of the ASAP dataset are presented in Table \ref{asap}. As previously mentioned, this dataset is composed of 8 subsets, each with unique prompts and scoring rubrics. Our Chinese Student English Essay (CSEE) dataset consists of 13,372 essays, along with their corresponding scores carefully rated by experienced English teachers based on the scoring standards in the Chinese National College Entrance Examination (Table \ref{rubrics}). The basic statistics of this dataset are outlined in Table \ref{private_stats}.

\begin{table}[h]
\centering
\small
\caption{Descriptive statistics of our private dataset. }
\label{private_stats}
\begin{tabular}{@{}lc@{}}
\toprule
\multicolumn{2}{l}{\textbf{Chinese Student English Essay Dataset}} \\ \midrule
\# of schools                         & 29                   \\
\# of essay prompts                       & 2               \\
\# of student essays                        & 13,372                 \\
avg. essay length                  & 124.74               \\
avg. Overall score                 & 10.72                \\
avg. Content score                 & 4.13                 \\
avg. Language score                & 4.05                 \\
avg. Structure score               & 2.55                 \\ \bottomrule
\end{tabular}
\end{table}

\begin{table*}[h]
\centering
\small
\caption{Descriptive statistics of the ASAP dataset. }
\label{asap}
\begin{tabular}{@{}clcccc@{}}
\toprule
Essay Set & \multicolumn{1}{c}{Essay Type}  & Grade Level & \# of Essays & Avg. Length & Score Range \\ \midrule
1         & Persuasive/Narrative/Expository & 8           & 1783         & 350         & {[}2, 12{]} \\
2         & Persuasive/Narrative/Expository & 10          & 1800         & 350         & {[}1, 6{]}  \\
3         & Source Dependent Responses      & 10          & 1726         & 150         & {[}0, 3{]}  \\
4         & Source Dependent Responses      & 10          & 1772         & 150         & {[}0, 3{]}  \\
5         & Source Dependent Responses      & 8           & 1805         & 150         & {[}0, 4{]}  \\
6         & Source Dependent Responses      & 10          & 1800         & 150         & {[}0, 4{]}  \\
7         & Persuasive/Narrative/Expository & 7           & 1569         & 300         & {[}0, 12{]} \\
8         & Persuasive/Narrative/Expository & 10          & 723          & 650         & {[}0, 36{]} \\ \bottomrule
\end{tabular}
\end{table*}

\section{Details of BERT Classifier Baseline}
\label{bert_appendix}

We employed the \textit{bert-base-uncased} BERT model from the huggingface transformers library\footnote{\url{https://huggingface.co/docs/transformers/}} using PyTorch. A simple fully connected layer was added to perform the classification task. The datasets were divided into training and testing sets at an 8:2 ratio. To ensure better reproducibility, we set all random seeds, including those for dataset splitting and model training, to the value 42. During training, we used cross-entropy loss as our loss function. We allowed BERT parameters to be fine-tuned, without freezing them, in line with the objective function. AdamW was chosen as the optimizer, with a learning rate set to $10^{-5}$ and epsilon at $10^{-6}$. With a batch size of 16 and a maximum of 10 training epochs, we also integrated an early stopping strategy to mitigate potential overfitting. All the experiments of the BERT baseline were run with 2 RTX A4000 16G GPUs in around one week.

\section{Details of LLM-based Methods}
\label{append_llm_methods}

\subsection{LLM Prompts}

The prompts used for LLMs in our study fall into two distinct categories: firstly, the zero-shot and few-shot configurations of GPT-4; secondly, the instructions for fine-tuning and inference of GPT-3.5 and LLaMA3-8B. The prompts for the few-shot scenario incorporate those used in the zero-shot setting and overlap with the fine-tuning prompts. Therefore, for clarity and conciseness, we present examples of the \textit{GPT-4, few-shot, with rubrics} and the inputs of fine-tuned LLaMA3-8B in Table \ref{example}.

\subsection{Few-Shot GPT-4} 

In the few-shot setting of GPT-4 with $k$ essay examples, as indicated by prior studies in AES tasks \citep{yancey-etal-2023-rating}, increasing the value of $k$ did not consistently yield better results, showing a trend of diminishing marginal returns. Therefore, we choose a suitable $k = 3$ in the study.

We explored two sampling approaches. The first involved randomly selecting essays from various levels of quality to help LLM understand the approximate level of the target essay. The second method adopted a retrieval-based approach, which has been proven to be effective in enhancing LLM performance \citep{Khandelwal2020Generalization, shi2023replug, ram2023context}. Leveraging OpenAI's \textit{text-embedding-ada-002} model, we calculated the embedding for each essay. This allowed us to identify the top $k$ similar essays based on cosine similarity (excluding the target essay). Our experiments demonstrated that this retrieval strategy consistently yielded superior results. Therefore, we focused on the latter approach in this study.

\subsection{Fine-tuning LLaMA3}

We fine-tuned the \textit{llama-3-8b-bnb-4bit} model using the \textit{unsloth} framework\footnote{\url{https://github.com/unslothai/unsloth}}. For this process, we employed a Parameter-Efficient Fine-Tuning (PEFT) approach with a rank of 16 and a LoRA alpha value of 16. We utilized an 8-bit AdamW optimizer, starting with an initial learning rate of $2 \times 10^{-4}$. After 50 warm-up steps, the learning rate was scheduled to decay linearly, with the weight decay parameter set at 0.01. We maintained all random seeds at 3407 and completed the fine-tuning over 2 epochs. All experiments involving the fine-tuned LLaMA3-8B were conducted using a single RTX 4090 24GB GPU, spanning approximately three weeks.

\begin{table*}[h]
\centering
\small
\caption{Generalization comparison of QWK scores for the Fine-tuned GPT-3.5 and the BERT Classifier under the ASAP dataset. }
\label{gen}
\begin{tabular}{@{}l|l|cccccccc@{}}
\toprule
                                  &                    & Set 1  & Set 2  & Set 3  & Set 4  & Set 5  & Set 6  & Set 7  & Set 8  \\ \midrule
\multirow{2}{*}{Trained on Set 1} & BERT Classifier      & -      & 0.3299 & 0.1680 & 0.1380 & 0.3045 & 0.1234 & 0.3002 & 0.1541 \\
                                  & Fine-tuned GPT-3.5 & -      & \textbf{0.5216} & \textbf{0.5405} & \textbf{0.4891} & \textbf{0.5076} & \textbf{0.6344} & \textbf{0.6306} & \textbf{0.3126} \\ \midrule
\multirow{2}{*}{Trained on Set 2} & BERT Classifier      & 0.2776 & -      & 0.1975 & 0.2392 & 0.1750 & 0.1453 & 0.2474 & 0.3783 \\
                                  & Fine-tuned GPT-3.5 & \textbf{0.4270} & -      & \textbf{0.4131} & \textbf{0.4619} & \textbf{0.5958} & \textbf{0.5579} & \textbf{0.5438} & \textbf{0.6684} \\ \midrule
\multirow{2}{*}{Trained on Set 3} & BERT Classifier      & 0.3468 & \textbf{0.4444} & -      & 0.6230 & 0.6319 & 0.5299 & 0.4368 & \textbf{0.2427} \\
                                  & Fine-tuned GPT-3.5 & \textbf{0.3991} & 0.2488 & -      & \textbf{0.7674} & \textbf{0.7714} & \textbf{0.7150} & \textbf{0.4964} & 0.1134 \\ \midrule
\multirow{2}{*}{Trained on Set 4} & BERT Classifier      & \textbf{0.3257} & \textbf{0.5332} & \textbf{0.6267} & -      & 0.5483 & 0.4959 & \textbf{0.4659} & 0.3204 \\
                                  & Fine-tuned GPT-3.5 & 0.0631 & 0.3493 & 0.4908 & -      & \textbf{0.6515} & \textbf{0.7420} & 0.0865 & \textbf{0.3419} \\ \midrule
\multirow{2}{*}{Trained on Set 5} & BERT Classifier      & 0.4051 & 0.3341 & 0.4264 & 0.4202 & -      & 0.5243 & \textbf{0.3255} & 0.2035 \\
                                  & Fine-tuned GPT-3.5 & \textbf{0.4354} & \textbf{0.4301} & \textbf{0.5765} & \textbf{0.6877} & -      & \textbf{0.7368} & 0.1061 & \textbf{0.3118} \\ \midrule
\multirow{2}{*}{Trained on Set 6} & BERT Classifier      & \textbf{0.3164} & 0.3462 & 0.4000 & 0.3067 & \textbf{0.4882} & -      & \textbf{0.2303} & \textbf{0.3047} \\
                                  & Fine-tuned GPT-3.5 & 0.1342 & \textbf{0.3607} & \textbf{0.4579} & \textbf{0.3157} & 0.3734 & -      & 0.0061 & 0.0859 \\ \midrule
\multirow{2}{*}{Trained on Set 7} & BERT Classifier      & 0.0975 & 0.0086 & 0.1854 & 0.0328 & 0.0554 & 0.1244 & -      & \textbf{0.2917} \\
                                  & Fine-tuned GPT-3.5 & \textbf{0.5862} & \textbf{0.3993} & \textbf{0.4865} & \textbf{0.4425} & \textbf{0.4494} & \textbf{0.4417} & -      & 0.2157 \\ \midrule
\multirow{2}{*}{Trained on Set 8} & BERT Classifier      & 0.0560 & 0.1102 & 0.0110 & 0.0164 & 0.0371 & 0.0454 & 0.1777 & -      \\
                                  & Fine-tuned GPT-3.5 & \textbf{0.2714} & \textbf{0.4822} & \textbf{0.4768} & \textbf{0.6009} & \textbf{0.4199} & \textbf{0.3231} & \textbf{0.5460} & -      \\ 
\bottomrule
\end{tabular}
\end{table*}

\section{Human-AI Co-Grading Details}

In our LLM-assisted human grading experiment, the 10 college students were all from a Normal University in Beijing, and had a male-to-female ratio of 4:6, with ages ranging from 19 to 23 years (from freshmen to seniors). Their English capabilities were certified by China's College English Test (CET). None of the novices have the experience of grading student essays currently. The 5 expert evaluators comprised experienced English teachers from Beijing high schools, with teaching tenures ranging from 8 to 20 years. Before evaluation, all participants received training on the standard scoring rubrics. They were also incentivized with appropriate remuneration for their participation.

The instructions for the evaluators include the standard scoring rubrics of the College Entrance Examination in China and several grading examples. The writing exercise and the essays designated for assessment will be presented to the evaluators. Moreover, supplementary feedback (scores, output confidences, and explanations) will be provided for the experimental groups. To enhance the evaluators' comprehension and avoid possible misunderstandings, all the information provided has been translated into Chinese. 

\begin{figure*}[h]
    \centering
    \includegraphics[width=0.48\textwidth]{img/total_res_2.png}
    \includegraphics[width=0.48\textwidth]{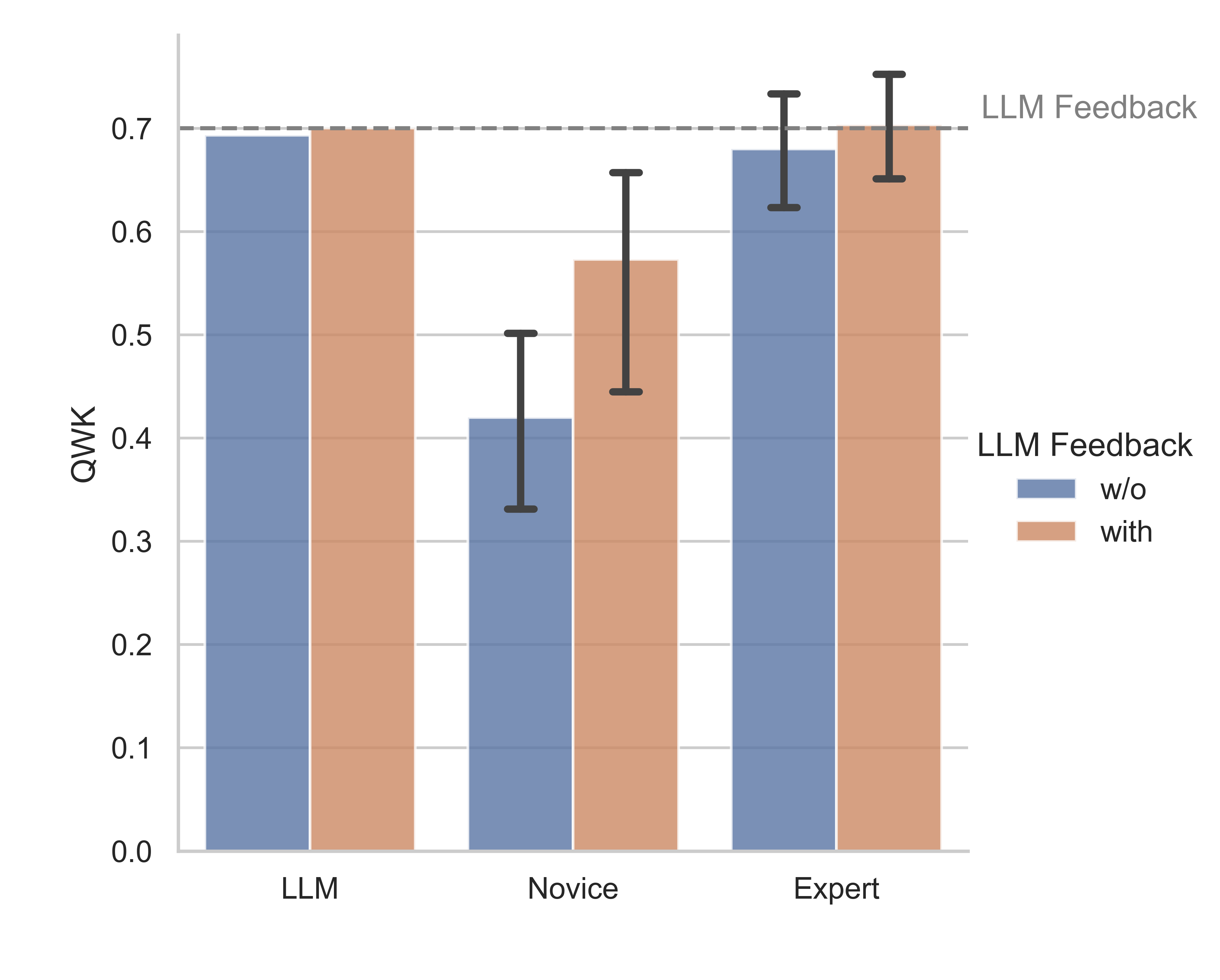}
    \includegraphics[width=0.48\textwidth]{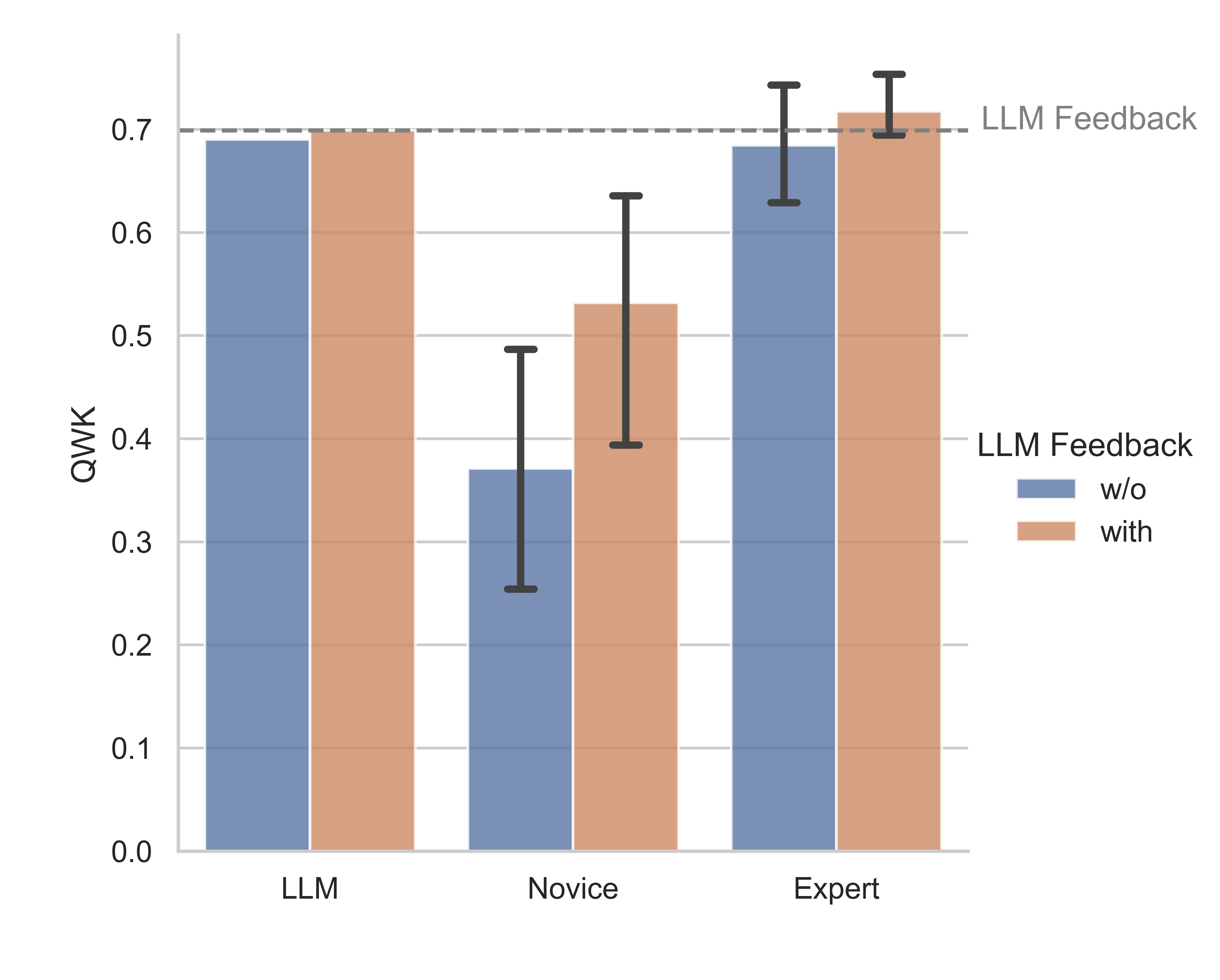}
    \includegraphics[width=0.48\textwidth]{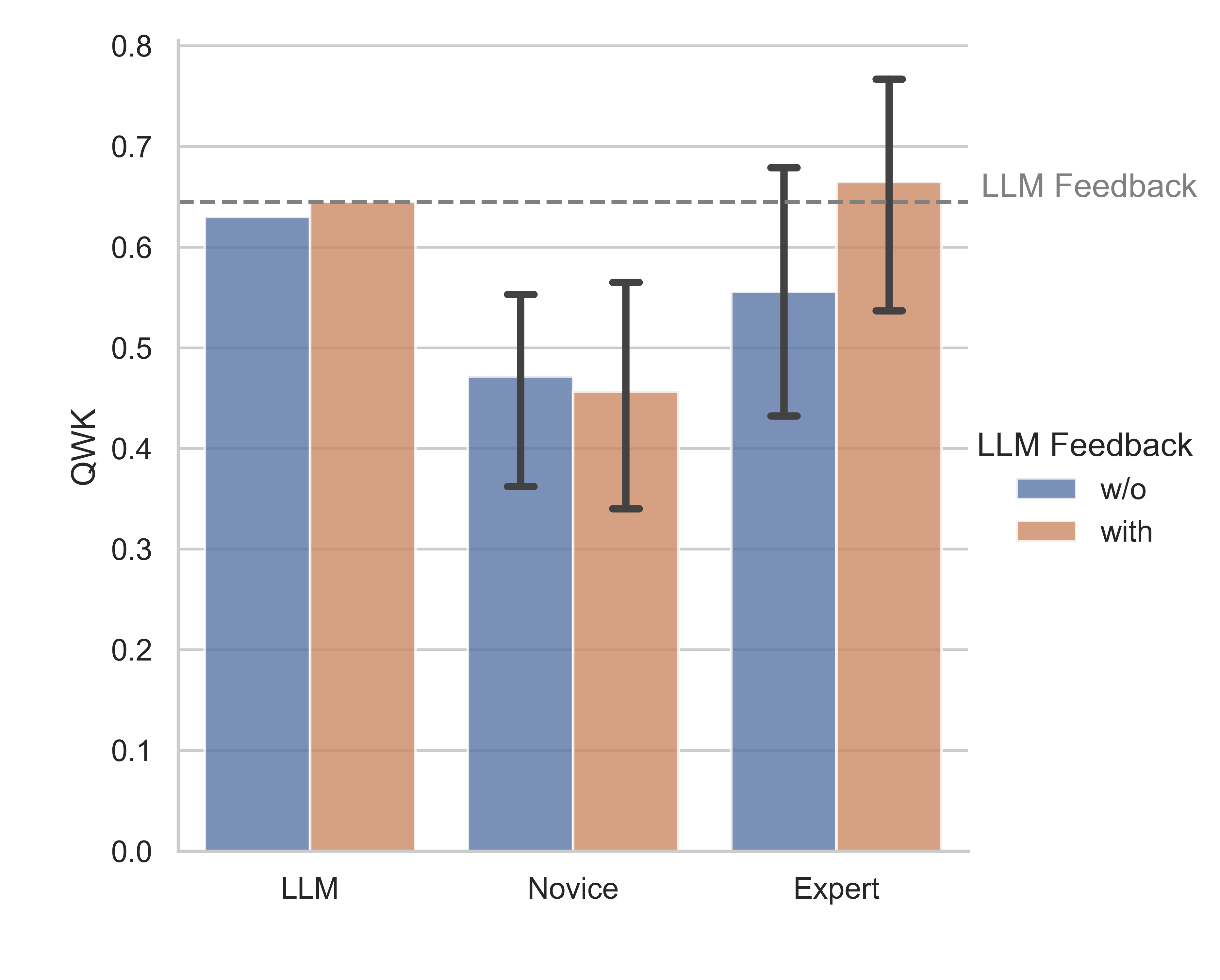}
    \caption{LLM-assisted grading experiment results for the novice, expert, and GPT-4 evaluators. From the top left to the bottom right is the result of the Overall score, Content score, Language score, and Structure score, respectively. }
    \label{detail_human_eval_res}
\end{figure*}

The results of Overall, Content, Language, and Structure scores in the human-AI co-grading experiment are presented in Figure \ref{detail_human_eval_res}. We observed that the Content and Language scores exhibit a similar trend as the Overall score discussed in the Results section. The expert evaluators noted that the Structure dimension is the most ambiguous and difficult part of the grading task which has the lowest QWK values among the three dimensions.

\begin{table*}[h]
\centering
\small
\caption{Rubrics for evaluating high school student essays in our private dataset. }
\label{rubrics}
\begin{tabular}{@{}p{0.9\textwidth}@{}}
\toprule
\multicolumn{1}{c}{\textbf{Rubrics}} \\ \midrule
\textbf{Overall Score} (20 points) = \textbf{Content Score} (8 points) + \textbf{Language Score} (8 points) + \textbf{Structure Score} (4 points)

\textbf{Content Dimension} (8 points in total)
\begin{itemize}
    \item 6-8 points:
    \begin{itemize}
        \item Content is complete with appropriate details
        \item Expression is closely related to the topic
    \end{itemize}
    \item 3-5 points:
    \begin{itemize}
        \item Content is mostly complete
        \item Expression is fundamentally related to the topic
    \end{itemize}
    \item 0-2 points:
    \begin{itemize}
        \item Content is incomplete
        \item Expression is barely related or completely unrelated to the topic
    \end{itemize}
\end{itemize}

\textbf{Language Dimension} (8 points in total)
\begin{itemize}
    \item 6-8 points:
    \begin{itemize}
        \item Language is accurate with diverse sentence structures and little or no errors (2 errors or fewer, 8 points; 3-4 errors, 7 points; 5-6 errors, 6 points)
        \item Language expression is mostly appropriate
    \end{itemize}
    \item 3-5 points:
    \begin{itemize}
        \item Language is not quite accurate, with some variation in sentence structures and several errors, but they don't impede understanding (7-8 errors, 5 points; 9-10 errors, 4 points; 11-12 errors, 3 points)
        \item Language expression is somewhat inappropriate
    \end{itemize}
    \item 0-2 points:
    \begin{itemize}
        \item Language is hopelessly inaccurate with numerous language errors, hindering understanding (more than 12 errors)
        \item Language expression is completely inappropriate
    \end{itemize}
\end{itemize}

\textbf{Structure Dimension} (4 points in total)
\begin{itemize}
    \item 3-4 points:
    \begin{itemize}
        \item Clearly and logically structured
        \item Smooth and coherent transitions
    \end{itemize}
    \item 1-2 points:
    \begin{itemize}
        \item Mostly clearly and logically structured
        \item Relatively smooth and coherent transitions
    \end{itemize}
    \item 0-1 points:
    \begin{itemize}
        \item Not clearly and logically structured
        \item Fragmented and disconnected structures and sentences
    \end{itemize}
\end{itemize}        \\ \bottomrule
\end{tabular}
\end{table*}

\begin{table*}[h]
\centering
\small
\caption{The example prompt (the \textit{GPT-4, few-shot, with rubrics} approach and the fine-tuned LLaMA3-8B) and the corresponding model output. }
\label{example}
\begin{tabular}{@{}p{0.6\textwidth}p{0.37\textwidth}@{}}
\toprule
\textbf{Example Prompt} & \textbf{Example Output} \\ \midrule
As a virtual evaluator with expertise in English composition, your role is to critically analyze and grade student essays according to a predetermined set of rubrics. You are to act as an impartial judge and evaluate the essays based on the quality of the writing and adherence to the essay prompt.







~\

Here are the specific guidelines for each score: 

\{\textit{Rubrics}\}

Sample Essay Prompt: 

\{\textit{Essay Prompt}\}

The graded example essays:

\{\textit{Examples}\}

Student's Essay to Evaluate: 

\{\textit{Target Student Essay}\}

~\

Task Breakdown:

1. Carefully read the provided essay prompt, scoring guidelines, and the student's essay.

2. In the Explanations part, identify specific elements in the essay referring to the rubrics. In the language dimension, list all the spelling and grammar errors, and count the number of them to determine the Language Score. The Explanations for each dimension should be as detailed as possible.

3. Determine the appropriate scores according to the analysis above. 

~\

Please present your evaluation in the following manner:

Explanations: ...,
Content Score: ...

Explanations: ...,
Language Score: ...

Explanations: ...,
Structure Score: ...

Explanations: ...,
Total Score: ...

Your final evaluation: 

[Total Score: ..., Content Score: ..., Language Score: ..., Structure Score: ...]               &  
Explanations: The student's essay responds accurately to Jim's request for suggestions on how to learn Chinese, providing thoughtful advice and support. Detailed suggestions include listening to Chinese radio, reading Chinese literature, and using mobile applications. The tone is hopeful and supportive.

Content Score: 8

~\

Explanations: There are no language errors; the grammar and spelling are all accurate. The student used formal, respectful language suitable for this kind of communication and also included diversity in sentence structures.

Language Score: 8

~\

Explanations: The essay is well-structured and organized. It follows the accepted structure of a reply email, with an opening greeting, a body containing the main content, and a closing remark. Transitions are also smooth and coherent, logically linking one thought to another.

Structure Score: 4

~\

Total Score:
This is the sum of the content, language, and structure scores.
Total Score: 20

~\

Your final evaluation:

[Total Score: 20, Content Score: 8, Language Score: 8, Structure Score: 4] \\ 
\midrule

Below is an instruction that describes a task, paired with an input that provides further context. Write a response that appropriately completes the request.

~\

\#\#\# Instruction:

You are an experienced English teacher analyzing high school students' essays according to a specific rubric. Evaluate the following essay based on three dimensions: Content, Language, and Structure, and provide the overall assessment. 

Please provide your evaluation in the following JSON format:

\{\textit{Output Format}\}

~\

\#\#\# Input:

Scoring rubrics:

\{\textit{Rubrics}\}

Essay Prompt:

\{\textit{Essay Prompt}\}

Student's Essay to Evaluate:

\{\textit{Target Student Essay}\}

~\

\#\#\# Response:

\{\}

& 

\{

'content':

\quad \{'completeness': 'The essay covers the basic requirements of the prompt by introducing design ideas for the graduation ceremony and explaining the reasons behind these designs. However, the explanations are quite brief and lack depth.', 

\quad 'topic\_relevance': 'The essay is related to the given topic, as it discusses ideas for a graduation ceremony design.', 

\quad 'content\_details': 'The details provided are minimal and lack depth. For example, the student mentions launching a ball in the dining hall and writing gratitude letters but does not elaborate on how these activities contribute to the overall design or the reasons behind these choices in detail.',

\quad 'score\_level': 'Level 2',

\quad 'score\_point': 4.5\}, 

'language': \{...\}, 

'structure': \{...\}, 

'overall': \{...\}

\}

\\ \bottomrule
\end{tabular}
\end{table*}

\end{document}